\pdfoutput=1

\documentclass[11pt]{article}

\usepackage[preprint]{acl}
\usepackage{times}
\usepackage{latexsym}

\usepackage[T1]{fontenc}

\usepackage[utf8]{inputenc}

\usepackage{microtype}

\usepackage{inconsolata}

\usepackage{graphicx}

\usepackage{xspace}
\usepackage{amsmath}
\usepackage{algorithm}
\usepackage{algpseudocode}
\usepackage{tabularx}
\usepackage{multirow}
\usepackage{makecell}
\usepackage{booktabs}
\usepackage{geometry}
\usepackage{enumitem}
\usepackage{tcolorbox}
\usepackage{tikz}
\usepackage{pgfplots}
\usepackage{pifont}
\usepackage{colortbl}
\definecolor{BBlue}{RGB}{91, 155, 213}
\definecolor{YYellow}{RGB}{255, 192, 0}
\definecolor{GGreen}{RGB}{112, 173, 71}
\definecolor{OOrange}{RGB}{237, 125, 49}
\definecolor{GGray}{RGB}{127, 127, 127}
\definecolor{PPink}{RGB}{255, 176, 195}

\title{Step-level Verifier-guided Hybrid Test-Time Scaling\\for Large Language Models}

\author{Kaiyan Chang\textsuperscript{\rm 1},
    Yonghao Shi\textsuperscript{\rm 1},
    Chenglong Wang\textsuperscript{\rm 1},
    Hang Zhou\textsuperscript{\rm 1},
    Chi Hu\textsuperscript{\rm 3}, \\
    \bf{Xiaoqian Liu}\textsuperscript{\rm 1},
    \bf{Yingfeng Luo}\textsuperscript{\rm 1},
    \bf{Yuan Ge}\textsuperscript{\rm 1},
    \bf{Tong Xiao}\textsuperscript{\rm 1,2}\thanks{\xspace Corresponding author.}, and
    \bf{Jingbo Zhu}\textsuperscript{\rm 1,2} \\
    \textsuperscript{\rm 1} School of Computer Science and Engineering, Northeastern University, Shenyang, China \\
    \textsuperscript{\rm 2} NiuTrans Research, Shenyang, China \\
    \textsuperscript{\rm 3} ByteDance, Hangzhou, China \\
    \ttfamily{changkaiyan\_neu@outlook.com},\\
    \ttfamily{\{xiaotong, zhujingbo\}@mail.neu.edu.cn}
}

\begin{document}

\maketitle
\begin{abstract}
Test-Time Scaling (TTS) is a promising approach to progressively elicit the model's intelligence during inference.
Recently, training-based TTS methods, such as continued reinforcement learning (RL), have further surged in popularity, while training-free TTS methods are gradually fading from prominence.
However, the additional computation overhead of training amplifies the burden on test-time scaling.
In this paper, we focus on training-free TTS methods for reasoning.
We first design \textit{Conditional Step-level Self-refinement}, a fine-grained sequential scaling method guided by process verification.
On top of its effectiveness, we further combine it with other classical parallel scaling methods at the step level, to introduce a novel inference paradigm called \textit{Hybrid Test-Time Scaling}\footnote{\url{https://github.com/Lucky-259/Hybrid_TTS}}.
Extensive experiments on five instruction-tuned LLMs across different scales (3B-14B) and families demonstrate that hybrid strategy incorporating various training-free TTS methods at a fine granularity has considerable potential for expanding the reasoning performance boundaries of LLMs.
\end{abstract}

\section{Introduction}
The remarkable breakthroughs of large language models (LLMs) on complex reasoning tasks~\cite{li2025system, xiao2025foundations, fan2025slam}, such as OpenAI-o1~\cite{openai_o1_system_card} and DeepSeek-R1~\cite{deepseekai2025deepseekr1incentivizingreasoningcapability}, are largely attributed to extensive inference-time computation for deep thinking, a paradigm known as ``Test-Time Scaling'' (TTS)~\cite{snell2024scaling}.
Current TTS methods can be broadly categorized into two groups based on training requirements:
Training-based methods continue training LLMs to generate longer chain-of-thoughts (CoT)~\cite{Wei2022ChainOT}, including reinforcement learning (RL)~\cite{Shao2024DeepSeekMathPT, yu2025dapo, zheng2025learning} and supervised fine-tuning (SFT)~\cite{morishita2024enhancing, muennighoff2025s1}.
Training-free methods guide LLMs to systematically explore the solution space, including parallel scaling~\cite{brown2024large, qimutual} and sequential scaling~\cite{gou2023critic, zhang2024small}. 

Although training-based TTS methods have shown outstanding reasoning performance, especially RL, they incur not only additional train-time computational cost but also unnecessary test-time computational overhead by ``overthinking'' simple problems~\cite{Chen2024DoNT}.
In contrast, training-free TTS methods can effectively mitigate these issues.
They typically control the generation process through verification~\cite{ji2025test, chen2025towards}, which guides the LLM toward correct scaling directions and stops scaling once a valid solution is confirmed.
Notably, the computational overhead of verification is significantly lower than that of generation, making verifier-guided TTS advantageous for its low operational cost and ``plug-and-play'' flexibility.
However, training-free TTS methods generally underperform on complex reasoning tasks, particularly evident in sequential scaling like self-refinement~\cite{Madaan2023SelfRefineIR, huanglarge} which yields marginal performance gains.

\begin{table}
    \centering
    \resizebox{0.99\columnwidth}{!}{%
    \begin{tabular}{lcc}
    \toprule[1.1pt]
    \multicolumn{1}{c}{\textbf{Test-Time Scaling Methods}}
    & \makecell{\textbf{Additional} \\ \textbf{Training}}
    & \makecell{\textbf{GPQA} \\ \textbf{Diamond}} 
    \\ \midrule
    Hybrid Test-Time Scaling (Ours) & \ding{53} & \textbf{51.5} \\
    Continued Reinforcement Learning & \ding{51} & 49.1 \\
    \bottomrule[1.1pt]
\end{tabular}}
    \caption{Reasoning performance activated by training-free and training-based TTS methods.}
    \label{tab_intro}
    \vspace{-1em}
\end{table}

In this paper, we explore how to effectively integrate various training-free TTS methods to further unlock the reasoning potential of LLMs.
To this end, we first propose \textit{Conditional Step-level Self-refinement} to validate the effectiveness of fine-grained sequential scaling.
On this basis, we further introduce \textit{Step-level Verifier-guided Hybrid Test-Time Scaling} as shown in Figure~\ref{fig_Hybrid_TTS}, which combines parallel (Best-of-N) and sequential (self-refinement) scaling within a step-level tree search.

Extensive experiments on MATH500, AIME24, and GPQA Diamond datasets show that Hybrid Test-Time Scaling consistently enhances the reasoning performance of five instruction-tuned LLMs across various scales (3-14B) and families (Qwen2.5, LLaMA3.1, Gemma3). Guided by the most capable process reward model (Qwen2.5-Math-PRM-7B), it achieves a maximum performance increase of 28.6\%.
Surprisingly, our method empowers the Qwen2.5-3B-Instruct model to surpass an RL-enhanced DeepSeek-R1-Distill-Qwen-7B model by 2.4\% on the GPQA Diamond dataset (Table~\ref{tab_intro}).
These findings suggest that the essence of test-time scaling lies in the extensive exploration and precise exploitation of the LLM's solution space.
Our primary contributions are as follows:
\begin{itemize}[leftmargin=*]
    \item To the best of our knowledge, we are the first to propose conditional step-level self-refinement and demonstrate the potential of fine-grained sequential scaling.
    \item We introduce Hybrid Test-Time Scaling, a novel inference paradigm that combines parallel and sequential scaling at the step level, leveraging a powerful verifier to efficiently navigate the model's solution space.
    \item We conduct a comprehensive empirical evaluation across diverse LLMs and challenging reasoning benchmarks. The consistent performance improvements not only illustrate the versatility and effectiveness of our hybrid strategy but also provide valuable insights into the integration of different training-free TTS methods.
\end{itemize}

\section{Related Works}
\subsection{Training-free Test-Time Scaling}
Training-free TTS methods fall into two categories: Parallel Scaling samples multiple responses in parallel and aggregates them into a final answer, while Sequential Scaling iteratively refines an initial response. 
Hybrid Scaling leverages the complementary benefits of these two paradigms, mirroring the human cognitive process of first generating diverse hypotheses (divergent thinking) and then systematically refining them toward an optimal solution (convergent thinking)~\cite{zhang2025and}.

Prior work has explored the integration of different training-free TTS methods. 
For instance, \citet{kang2024mindstar, liucan, wu2024scaling} introduce the core idea of Best-of-N into the search process to improve inference efficiency. However, their explorations remain confined to the parallel scaling.
Other research explores hybrid scaling, typically integrating sequential scaling into a structured search framework to achieve the exploration-exploitation trade-off in the solution space:

\begin{itemize}[leftmargin=*]
    \item \textbf{Solution-level.} One line of work considers a complete solution as a node in Monte Carlo Tree Search (MCTS)~\cite{zhang2024accessing, rabby2024mc, zhang2025llama}, where iterative self-refinement acts as the simulation process. Nevertheless, such solution-level exploration remains insufficient for complex reasoning.
    \item \textbf{Step-level.} Another line of work treats each reasoning step as a node. Representative methods like Tree-of-Thought (ToT)~\cite{yao2023tree} and Graph-of-Thought (GoT)~\cite{besta2024graph} construct a network (tree or arbitrary graph) of thoughts and prompt the LLM itself to evaluate and refine incorrect steps. SR-MCTS~\cite{zhao2024marco} further refines this to mini-steps (32/64 tokens), using the log probabilities of alternative tokens to evaluate node values. Despite the finer granularity of step-level exploration, the self-feedback mechanism is susceptible to the model's inherent hallucinations, thereby limiting its overall reasoning performance.
\end{itemize}

Therefore, we aim to introduce a high-quality step-level verifier to provide precise and real-time evaluations of intermediate reasoning steps, guiding the LLM to prevent error accumulation and scale toward the correct solution efficiently.

\subsection{Verifiers}\label{sec:Verifiers}
The verification mechanism plays a crucial role in TTS. It can be used to select the optimal solution from candidate responses in parallel scaling and determine when to stop generation in sequential scaling~\cite{zhang2025and}.
One of the most common forms of verification is the LLM-based generative reward model~\cite{zhanggenerative, wanggram}.
Two types of verifiers have emerged based on whether they evaluate the final answer or the reasoning path: outcome reward models (ORMs) and process reward models (PRMs).

To investigate the efficacy of fine-grained TTS, we focus on process-level verifiers capable of accurately evaluating intermediate reasoning steps.
The concept of process verification was first introduced by DeepMind~\cite{uesato2022solving}, laying the foundation for subsequent research. 
OpenAI~\cite{lightman2023let} later trains a reward model on a larger-scale, human-annotated dataset PRM800K, demonstrating that PRMs outperform ORMs in evaluating reasoning quality.
Since then, research on PRM training has become a prominent area of research. 
For instance, Math-Shepherd~\cite{wang2023math} employs automatically constructed process-wise supervision data instead of manual annotation.
\citet{zhang2025lessons} propose a consensus filtering mechanism that combines MC estimation with LLM-as-a-judge to improve both model performance and data efficiency in PRM training.

In this paper, the verifier's feedback dictates the scaling direction of each TTS method. 
Therefore, it is natural to select a verifier capable of providing accurate step-level verification.
According to PRMBench~\cite{song2025prmbench}, a comprehensive benchmark for PRMs, Qwen2.5-Math-PRM-7B~\cite{zhang2025lessons} is currently the most capable PRM among those with 7B/8B parameters.
Thus, we adopt it as our step-level verifier to investigate fine-grained Hybrid Test-Time Scaling.

\begin{figure*}
    \centering
    \includegraphics[width=0.99\linewidth]{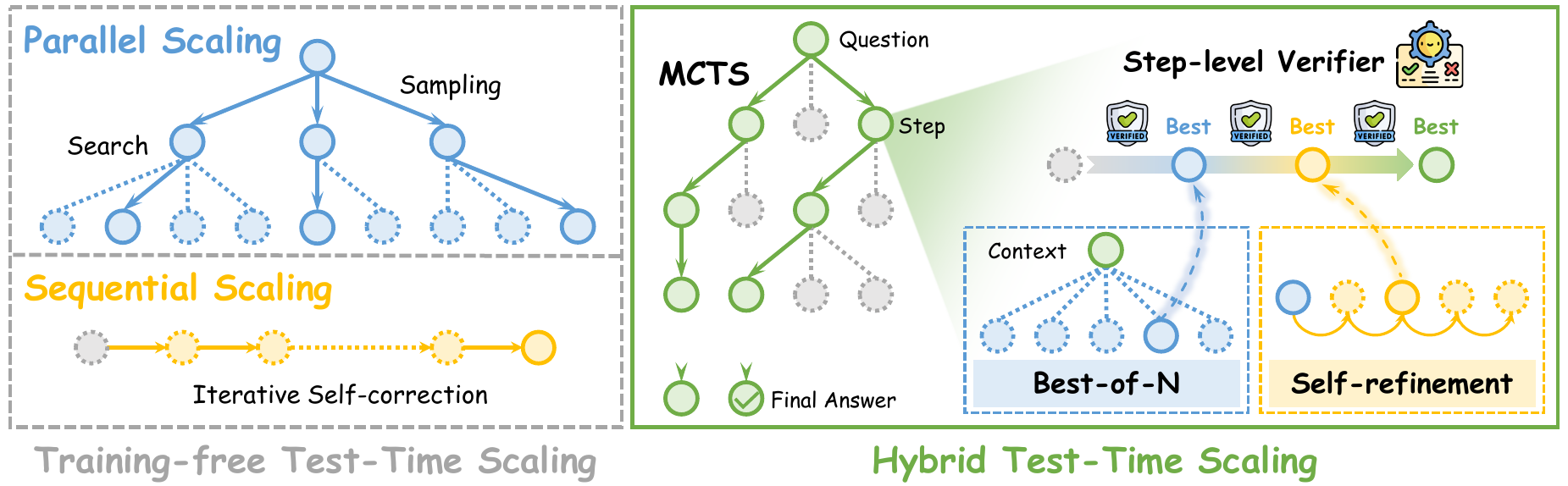}
    \caption{\textbf{Left}: The two mainstream paradigms of training-free TTS, parallel and sequential scaling. \textbf{Right}: Our proposed Step-level Verifier-guided Hybrid Test-Time Scaling, which incorporates representative methods from both paradigms. Specifically, it performs ``Deep Scaling'' on each reasoning step within an MCTS framework through a two-stage screening process: (1) Multiple candidates are generated via Best-of-N sampling and the best step is selected using the PUCT algorithm. (2) This candidate undergoes a PRM-guided conditional self-refinement to produce the final best step, completing the ``\textit{best-of-the-best}'' selection cycle.}
    \label{fig_Hybrid_TTS}
    \vspace{-1em}
\end{figure*}

\section{Method}

\subsection{Control of Generation Granularity}
Controlling the LLM generation granularity at the step level is a precondition for fine-grained training-free TTS.
We observe that instruction-tuned open-source LLMs (\emph{e.g.} Qwen, LLaMA, and Gemma) naturally segment reasoning steps using ``\verb|\n|\verb|\n|'' when performing reasoning tasks.
We thus implement a ``Pause-then-Continue'' strategy to control their generation granularity.
Specifically, we set ``\verb|\n|\verb|\n|'' as a delimiter that signals the model to pause after generating each reasoning step. We then append ``\verb|\n|'' to the context to prompt the model to resume generation until a terminator token ``\verb|<eos>|''. 
This ``Pause-then-Continue'' strategy requires LLMs to have strong instruction-following and reasoning capabilities.
Therefore, we utilize instruction-tuned LLMs with at least 3B parameters to investigate fine-grained TTS.

\subsection{Conditional Step-level Self-refinement}
Self-refinement is a representative sequential scaling method, where meta-prompts guide the LLM to automatically perform feedback and revision on the initial solution.
However, its improvements in reasoning performance are significantly lower than other common parallel scaling methods.
We hypothesize that the scaling granularity of solution-level self-refinement is too coarse-grained, leading to inaccurate identification of erroneous steps or unnecessary modifications of correct ones.
Therefore, we propose a conditional step-level self-refinement method for fine-grained sequential scaling, where the verification process plays a critical role.

\paragraph{Why verify?} 
We focus on self-refinement for reasoning tasks. The LLM first generates a solution based on a CoT prompt, then reflects on the weaknesses with a Critic prompt, and finally revises the original solution based on the question, current solution, and reflection guided by a Rewrite prompt.
In fact, the above reflection and revision process is often conducted unconditionally, which introduces two major concerns.
Firstly, forcing self-reflection even when the solution is already correct can cause the model to second-guess its own valid reasoning, introducing the risk of erroneously correcting a valid answer and incurring unnecessary refinement overhead.
Secondly, in the absence of a clear stopping signal, the LLM may overlook an already optimal solution and continue making unnecessary or even detrimental modifications.
Hence, it is crucial to ensure the correctness of the reasoning process through accurate and real-time verification for superior performance.

\paragraph{What to verify?} 
Step-by-step CoT reasoning is sequential and prone to error accumulation, as early-stage errors can often lead to incorrect final answers.
Therefore, we aim to verify the correctness of the logical relationship between the current step and the preceding context, thereby ensuring the quality of intermediate reasoning steps.
Importantly, the verification of the current step takes place before the generation of the next step, a design that yields two benefits: (i) it ensures subsequent steps are generated based on verified high-quality context, thereby mitigating the typical error accumulation in long CoT; and (ii) it prevents unnecessary refinement of already correct steps, reducing the likelihood of over-refinement or hallucination-induced overthinking.

\paragraph{How to verify?} 
The PRM is a fine-grained verifier that scores each reasoning step on a scale of 0 to 1 based on the question and preceding steps.
We thus leverage the reliable evaluation capabilities of the PRM to design our conditional step-level self-refinement strategy with the following principles:

\begin{itemize}[leftmargin=*]
    \item \textbf{Whether to reflect.} Reflection is triggered when the PRM score for the current step falls below a predefined threshold, indicating low confidence in its correctness.
    \item \textbf{How to update.} The reasoning step is updated only if the PRM score of the revised step exceeds that of the current step; otherwise, the current step is retained.
    \item \textbf{When to stop.} Refinement stops when the PRM score surpasses a target threshold, when incremental improvements plateau, or when a maximum number of iterations is reached.
\end{itemize}

A key property of our proposed conditional step-level self-refinement is its orthogonality to other training-free TTS methods. To further balance performance and efficiency, we conducted a series of preliminary experiments to identify the optimal conditions for improved step-level self-refinement, as discussed in Section~\ref{sec:preliminary_experiments}.

\subsection{Step-level Verifier-guided Hybrid Test-Time Scaling}
Inspired by the effectiveness of step-level self-refinement, we further investigate the potential of training-free test-time scaling at a finer granularity.
Specifically, we aim to integrate representative techniques in both parallel and sequential scaling within a unified step-level framework.
As illustrated on the right side of Figure~\ref{fig_Hybrid_TTS}, our Hybrid Test-Time Scaling paradigm synergistically combines Best-of-N (BoN) and self-refinement in a tree search framework at the step level, all guided by a PRM that serves as the step-level verifier to ensure the quality of each intermediate reasoning step.

In concrete terms, we construct a reasoning tree where the initial question serves as the root node, and each subsequent node represents a reasoning step.
Our algorithm~\ref{al:Hybrid_Scaling} for step-level Hybrid Test-time Scaling (detailed in Appendix~\ref{appendix:algorithm}) employs the PRM to evaluate the value of each candidate step.
The most promising step with the highest score is then selected for expansion as the next node in the reasoning path.
In this way, an optimal reasoning path is progressively constructed, ultimately forming a complete solution.
The three-stage pipeline for scaling each step is as follows: 

\begin{enumerate}[leftmargin=*]
    \item \textbf{Best-of-N.} At each step of the tree search, we begin from the current node and perform N parallel samplings of candidate reasoning steps. To balance exploration and exploitation, we select the best action $a$ by maximizing a score adapted from the Probabilistic Upper Confidence Tree (PUCT)~\cite{schrittwieser2020mastering} as defined in Eq.~\ref{eq:PUCT}. This score integrates the value $Q(s, a)$ estimated by a PRM, the prior probability $P(s, a)$ given by the LLM, and the visit count $N(s, a)$.

    \begin{equation}
    \begin{split}
    \label{eq:PUCT}
    & \arg\max_{a}\bigg[Q(s, a) + P(s, a) \frac{\sqrt{N(s, b)}}{1 + N(s, a)} \\
    & \cdot\bigg(c_1 + \log\Big(\frac{N(s, b) + c_2 + 1}{c_2}\Big)\bigg) \bigg]
    \end{split}
    \end{equation}
    
    where $s$ denotes the current state, \emph{i.e.}, the parent node, which includes the question and all previously generated reasoning steps. $a$ represents a specific candidate action from state $s$, \emph{i.e.}, a possible next reasoning step to be generated. $N(s, b)$ corresponds to the total number of visits to the parent node $s$. The hyperparameters $c_1$ and $c_2$ are used to control the influence of the policy term relative to the value term (exploration degree). In our experiments, we set $c_1$ = 1.25 and $c_2$ = 19,652.
    \item \textbf{Conditional Self-refinement.} The winning candidate from the Best-of-N stage then undergoes a conditional self-refinement process. Specifically, self-refinement is triggered only if its PRM score falls below a predefined threshold. In such cases, a revised step is generated based on the self-feedback, and adopted only if its PRM score exceeds that of the original step; otherwise, the original step is retained.
    \item \textbf{Monte Carlo Tree Search.} Finally, the optimal candidate step that passes both the Best-of-N and conditional self-refinement stages is officially selected as the final step and appended to the current search trajectory. It then becomes the new leaf node from which the subsequent search iteration will expand.
\end{enumerate}

We conceptualize this pipeline as ``Deep Scaling'' applied to the simulation phase, which is the most critical component of the MCTS.
In a standard MCTS, the simulation (or ``rollout'') typically starts from the current reasoning step and generates the remaining part of the solution with a default policy. 
Subsequently, an evaluation score based on the correctness of the final answer is back-propagated to update the value estimates of all nodes along the reasoning path.
In contrast, ``Deep Scaling'' reshapes the above process.
Since the PRM can provide immediate and precise value assessments for each intermediate step, it eliminates the need to complete the entire solution before evaluating node quality. 
As a result, we focus computational resources on the deep scaling of each intermediate step, performing a more deliberate exploration through multi-sampling and iterative refinement.
In summary, the introduction of the step-level verifier provides accurate and efficient guidance for LLMs during test-time scaling.
Meanwhile, the integration of multiple training-free TTS strategies ensures both the breadth and depth of exploration in the solution space, achieving a comprehensive performance improvement.

\section{Experimental Setup}
\paragraph{Models.} For LLMs, we include Qwen2.5-3B-Instruct, Qwen2.5-7B-Instruct, Qwen2.5-14B-Instruct ~\cite{yang2024qwen2}, LLaMA3.1-8B-Instruct~\cite{grattafiori2024llama}, and Gemma3-4B-IT~\cite{team2025gemma} to span a range of model scales. We deliberately exclude RL-enhanced LLMs because their implicit CoT reasoning prior to the final answer lacks a clear step-by-step structure. Moreover, we aim to explore how to effectively activate the inherent reasoning potential of instruction-tuned LLMs, and provide insights for training-based TTS methods. For PRMs, we select Qwen2.5-Math-PRM-7B~\cite{zhang2025lessons} as the step-level verifier for process verification, for reasons detailed in Section \ref{sec:Verifiers}.

\paragraph{Datasets.} We mainly select three mathematical and reasoning datasets: MATH500~\cite{hendrycks2021measuring}, AIME24~\cite{AIME}, and GPQA Diamond~\cite{rein2024gpqa}. They span a range of domains including mathematical computation, complex reasoning, and scientific questions, and present a significant difficulty level.

\paragraph{Implementation.} We implement the Hybrid Test-Time Scaling method based on OpenR\footnote{\url{https://openreasoner.github.io}}, an open-source framework designed to integrate key components for enhancing the reasoning capabilities of LLMs. All experiments are run once on two 80G A800 GPUs based on vLLM~\cite{kwon2023efficient} framework for efficient LLM inference, with one GPU deploying LLM service and another for PRM service. All hyperparameters are consistent with the model's default generation configuration.

\paragraph{Evaluation.} We evaluate the reasoning ability of LLMs with three metrics: 
\begin{itemize}[leftmargin=*]
    \item \textbf{Maj@k} measures the majority vote correctness from $k$ independently sampled outputs based on self-consistency~\cite{wang2022self}.
    \item \textbf{RM@k} measures the best solution correctness from $k$ sampled outputs selected by the specific reward model. Our strategy is to select the reasoning path that exhibits the maximum of per-path minimum PRM scores as the final solution.
    \item \textbf{Pass@k} measures whether at least one of the model's $k$ sampled outputs is correct.
\end{itemize}

\paragraph{Baselines.} We compare the conditional step-level self-refinement method against other parallel scaling methods and solution-level self-refinement under identical verifier and iteration settings. As for hybrid test-time scaling, it incorporates MCTS, BoN, and step-level self-refinement; we thus regard pairwise combinations as comparative baselines. Additionally, we test solution-level hybrid scaling methods (OpenR) proposed by~\citet{wang2410openr} under equivalent sampling, search path, and refinement conditions.

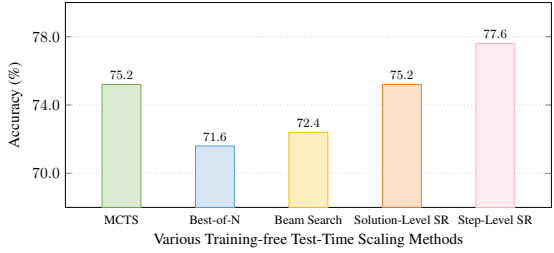
\begin{figure}
    \centering
    \resizebox{0.95\linewidth}{!}{
    \begin{tikzpicture}
    \begin{axis}[
        at={(0,0)},
        anchor=south west,
        height=0.27\textheight,
        width=0.85\textwidth,
        ymajorgrids,
        grid style=dotted,
        ybar,
        bar width=2.5em,
        bar shift=0pt,
        enlarge x limits=0.15,
        nodes near coords,
        nodes near coords align={vertical},
        nodes near coords style={font=\small,/pgf/number format/fixed, /pgf/number format/fixed zerofill, /pgf/number format/precision=1},
        ylabel={Accuracy (\%)},
        xlabel={Various Training-free Test-Time Scaling Methods},
        symbolic x coords={MCTS, Best-of-N, Beam Search, Solution-Level SR, Step-Level SR},
        xtick={MCTS, Best-of-N, Beam Search, Solution-Level SR, Step-Level SR},
        x tick label style={font=\small,inner sep=1pt},
        xtick style={draw=none},
        ymin=68, ymax=80,
        ytick={70,74,78},
        yticklabel style={/pgf/number format/fixed, /pgf/number format/fixed zerofill, /pgf/number format/precision=1},
    ]
        \addplot [fill=GGreen!25, draw=GGreen] coordinates {(MCTS, 75.2)};
        \addplot [fill=BBlue!25, draw=BBlue] coordinates {(Best-of-N, 71.6)};
        \addplot [fill=YYellow!25, draw=YYellow] coordinates {(Beam Search, 72.4)};
        \addplot [fill=OOrange!25, draw=OOrange] coordinates {(Solution-Level SR, 75.2)};
        \addplot [fill=PPink!25, draw=PPink] coordinates {(Step-Level SR, 77.6)};
    \end{axis}
\end{tikzpicture}}
    \caption{Comparisons between four common training-free TTS methods with Qwen2.5-7B-Instruct on the MATH500 dataset, where ``SR'' represents the basic self-refinement for one iteration.}
    \label{fig_comparisons}
    \vspace{-0.5em}
\end{figure}

\section{Results and Analyses}
\subsection{Preliminary Experiments for Conditional Step-level Self-refinement}\label{sec:preliminary_experiments}
\paragraph{Basic step-level self-refinement outperforms other training-free test-time scaling methods.} 
We first evaluated both solution-level and step-level self-refinement, each for one iteration, using the prompt templates detailed in Appendix~\ref{appendix:meta-prompt}.
As shown in Figure~\ref{fig_comparisons}, our basic step-level self-refinement achieves superior reasoning performance compared to other training-free TTS baselines reported in \citet{ding2025dynamic}.

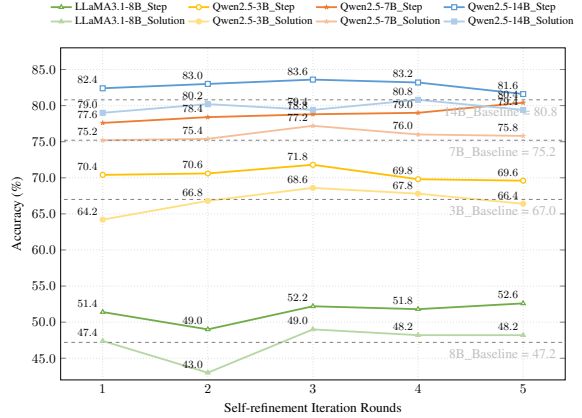
\begin{figure}
    \centering
    \resizebox{0.99\linewidth}{!}{
    \begin{tikzpicture}
\begin{axis}[
    at={(0,0)},
    anchor=south west,
    height=0.45\textheight,
    width=\textwidth,
    grid=major,
    major grid style={dotted},
    enlarge x limits=0.1,
    nodes near coords,
    nodes near coords align={south east},
    nodes near coords style={text=black, font=\small, /pgf/number format/fixed, /pgf/number format/fixed zerofill, /pgf/number format/precision=1},
    ymin=42, ymax=88,
    yticklabel style={/pgf/number format/fixed, /pgf/number format/fixed zerofill, /pgf/number format/precision=1},
    ylabel={Accuracy (\%)},
    ylabel style={font=\normalsize, anchor=center},
    xtick style={font=\normalsize, draw=none},
        legend entries={LLaMA3.1-8B\_Step,Qwen2.5-3B\_Step,Qwen2.5-7B\_Step,Qwen2.5-14B\_Step,LLaMA3.1-8B\_Solution,Qwen2.5-3B\_Solution,Qwen2.5-7B\_Solution,Qwen2.5-14B\_Solution},
    legend columns=4,
    legend style={
        at={(axis cs:3,95)},
        anchor=north,
        draw=none,
        legend cell align=left,
        font=\small,
    },
    xmin=1, xmax=5,
    xtick={1,2,3,4,5},
    xlabel={Self-refinement Iteration Rounds},
]
    \addplot [sharp plot, GGreen, line width=1.5pt, mark=triangle*, mark size=2pt, mark options={fill=white,draw=GGreen,line width=1pt}]
    coordinates {
    (1,51.4)(2,49.0)(3,52.2)(4,51.8)(5,52.6) 
    };
    \addplot [sharp plot, YYellow, line width=1.5pt, mark=*, mark size=2pt, mark options={fill=white,draw=YYellow,line width=1pt}]
    coordinates {
    (1,70.4)(2,70.6)(3,71.8)(4,69.8)(5,69.6)
    };
    \addplot [sharp plot, OOrange, line width=1.5pt, mark=star, mark size=2pt, mark options={fill=white,draw=OOrange,line width=1pt}]
    coordinates {
    (1,77.6)(2,78.4)(3,78.8)(4,79.0)(5,80.4)
    };
    \addplot [sharp plot, BBlue, line width=1.5pt, mark=square*, mark size=2pt, mark options={fill=white,draw=BBlue,line width=1pt}]
    coordinates {
    (1,82.4)(2,83.0)(3,83.6)(4,83.2)(5,81.6)
    };
    \addplot [sharp plot, GGreen!50, line width=1.5pt, mark=triangle*, mark size=2pt, mark options={fill=GGreen!50,draw=GGreen!50,line width=1pt}]
    coordinates {
    (1,47.4)(2,43.0)(3,49.0)(4,48.2)(5,48.2)
    };
    \addplot [sharp plot, YYellow!50, line width=1.5pt, mark=*, mark size=2pt, mark options={fill=YYellow!50,draw=YYellow!50,line width=1pt}]
    coordinates {
    (1,64.2)(2,66.8)(3,68.6)(4,67.8)(5,66.4)
    };
    \addplot [sharp plot, OOrange!50, line width=1.5pt, mark=star, mark size=2pt, mark options={fill=OOrange!50,draw=OOrange!50,line width=1pt}]
    coordinates {
    (1,75.2)(2,75.4)(3,77.2)(4,76.0)(5,75.8)
    };
    \addplot [sharp plot, BBlue!50, line width=1.5pt, mark=square*, mark size=2pt, mark options={fill=BBlue!50,draw=BBlue!50,line width=1pt}]
    coordinates {
    (1,79.0)(2,80.2)(3,79.4)(4,80.8)(5,79.4)
    };
    \draw[dashed, gray, thick] (axis cs:0,47.2) -- (axis cs:6,47.2);
    \node[anchor=west, font=\normalsize, color=gray!50] at (axis cs:4.7,47.2) [yshift=-10pt, xshift=-3.5em] {8B\_Baseline = 47.2};
    \draw[dashed, gray, thick] (axis cs:0,67.0) -- (axis cs:6,67.0);
    \node[anchor=west, font=\normalsize, color=gray!50] at (axis cs:4.7,67.0) [yshift=-10pt, xshift=-3.5em] {3B\_Baseline = 67.0};
    \draw[dashed, gray, thick] (axis cs:0,75.2) -- (axis cs:6,75.2);
    \node[anchor=west, font=\normalsize, color=gray!50] at (axis cs:4.7,75.2) [yshift=-10pt, xshift=-3.5em] {7B\_Baseline = 75.2};
    \draw[dashed, gray, thick] (axis cs:0,80.8) -- (axis cs:6,80.8);
    \node[anchor=west, font=\normalsize, color=gray!50] at (axis cs:4.65,80.8) [yshift=-10pt, xshift=-3.5em] {14B\_Baseline = 80.8};
\end{axis}
\end{tikzpicture}}
    \caption{Multi-round iterative self-refinement at the solution-level and step-level on the MATH500 dataset.}
    \label{fig_ssself-refine}
    \vspace{-0.5em}
\end{figure}

\paragraph{Sequential scaling at the step level is superior to solution level.}
Since self-refinement achieves sequential scaling through multi-round iterative optimization, we further increased the number of refinement iterations.
We employed Qwen2.5-Math-PRM-7B as a verifier to conditionally accept a refined response only if it improved upon the original. Specifically, the solution quality was determined by the PRM score of its final answer. As presented in Figure~\ref{fig_ssself-refine}, we observe that step-level self-refinement consistently outperforms solution-level across four different LLMs, revealing considerable advantages of fine-grained sequential scaling.

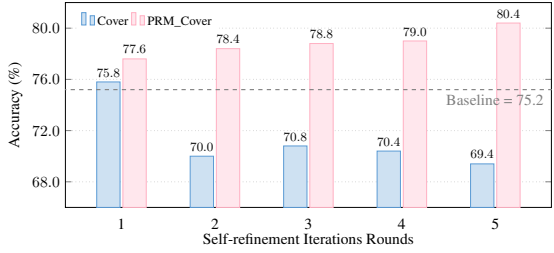
\begin{figure}
    \centering
    \resizebox{0.95\linewidth}{!}{
    \begin{tikzpicture}
    \begin{axis}[
        at={(0,0)},
        anchor=south west,
        height=0.27\textheight,
        width=0.85\textwidth,
        ymajorgrids,
        grid style=dotted,
        ybar,
        bar width=1.5em,
        enlarge x limits=0.15,
        nodes near coords,
        nodes near coords align={vertical},
        nodes near coords style={font=\small,/pgf/number format/fixed, /pgf/number format/fixed zerofill, /pgf/number format/precision=1},
        ylabel={Accuracy (\%)},
        xlabel={Self-refinement Iterations Rounds},
        xmin=1, xmax=5,
        xtick=data,
        xticklabels={1,2,3,4,5},
        xtick style={draw=none},
        ymin=66, ymax=82,
        ytick={68,72,76,80},
        yticklabel style={/pgf/number format/fixed, /pgf/number format/fixed zerofill, /pgf/number format/precision=1},
        legend entries={Cover, PRM\_Cover},
        legend style={
            at={(0.17,0.85)},
            anchor=south,
            legend columns=-1,
            font=\small,
            draw=none,
        },
    ]
    \addplot[fill=BBlue!30, draw=BBlue] coordinates {
        (1,75.8) (2,70.0) (3,70.8) (4,70.4) (5,69.4)
    };
    \addplot[fill=PPink!30, draw=PPink] coordinates {
        (1,77.6) (2,78.4) (3,78.8) (4,79.0) (5,80.4)
    };
    \draw[dashed, gray, thick] (axis cs:0,75.2) -- (axis cs:6,75.2);
    \node[anchor=west, font=\normalsize, color=gray] at (axis cs:5,75.2) [yshift=-7pt, xshift=-3.5em] {Baseline = 75.2};
    \end{axis}
\end{tikzpicture}}
    \caption{Experiments on different update strategies.}
    \label{fig_condition_cover}
    \vspace{-0.5em}
\end{figure}

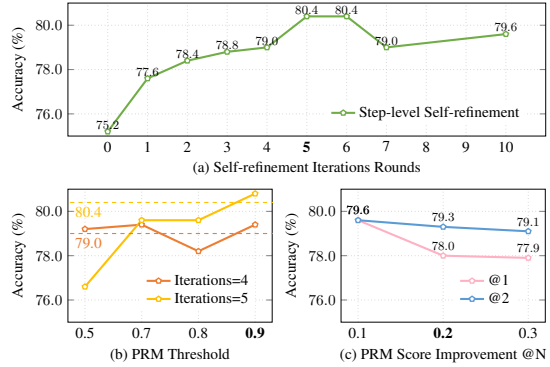
\begin{figure}
    \centering
    \resizebox{0.95\linewidth}{!}{
    \pgfplotsset{
    myaxis style/.style={
        height=0.2\textheight,
        grid=major,
        major grid style={dotted},
        enlarge x limits=0.10,
        ymin=75, ymax=81,
        ytick={76,78,80},
        yticklabel style={/pgf/number format/fixed, /pgf/number format/fixed zerofill, /pgf/number format/precision=1},
        ylabel={Accuracy (\%)},
        ylabel style={font=\normalsize, anchor=center},
        xtick style={font=\normalsize, draw=none},
    },
}

\begin{tikzpicture}
\begin{axis}[
    myaxis style,
    name=plot1,
    at={(0,0)},
    anchor=south west,
    width=0.85\textwidth,
    nodes near coords,
    nodes near coords align={south},
    nodes near coords style={text=black, font=\small, yshift=-0.2em, /pgf/number format/fixed, /pgf/number format/fixed zerofill, /pgf/number format/precision=1},
    legend entries={Step-level Self-refinement},
    legend style={
        at={(axis cs:8,75.5)},
        anchor=south,
        draw=none,
    },
    xmin=0, xmax=10,
    xtick={0,1,2,3,4,5,6,7,8,9,10},
    xticklabels={0,1,2,3,4,\textbf{5},6,7,8,9,10},
    xlabel={(a) Self-refinement Iterations Rounds},
]
    \addplot [sharp plot, GGreen, line width=1.5pt, mark=pentagon*, mark size=2pt, mark options={fill=white,draw=GGreen,line width=1pt}]
    coordinates {
    (0,75.2)(1,77.6)(2,78.4)(3,78.8)(4,79.0)(5,80.4)(6,80.4)(7,79.0)(10,79.6)
    };
\end{axis}

\newlength{\plotVerticalGap}
\setlength{\plotVerticalGap}{-3.5em}

\newlength{\plotHorizontalGap}
\setlength{\plotHorizontalGap}{4.5em}

\begin{axis}[
    myaxis style,
    name=plot2,
    at={(plot1.south west)},
    anchor=north west,
    yshift=\plotVerticalGap,
    width=0.42\textwidth,
    legend style={
        at={(axis cs:3,75.5)},
        anchor=south,
        draw=none,
        legend columns=1,
    },
    xmin=1, xmax=4,
    xtick={1,2,3,4},
    xticklabels={0.5,0.7,0.8,\textbf{0.9}},
    xlabel={(b) PRM Threshold},
]
    \addplot [sharp plot, OOrange, line width=1.5pt, mark=pentagon*, mark size=2pt, mark options={fill=white,draw=OOrange,line width=1pt}]
    coordinates {
    (1,79.2)(2,79.4)(3,78.2)(4,79.4)
    };
    \addlegendentry{Iterations=4}

    \addplot [sharp plot, YYellow, line width=1.5pt, mark=pentagon*, mark size=2pt, mark options={fill=white,draw=YYellow,line width=1pt}]
    coordinates {
    (1,76.6)(2,79.6)(3,79.6)(4,80.8)
    };
    \addlegendentry{Iterations=5}

    \draw[dashed, OOrange, thick] (axis cs:0,79.0) -- (axis cs:5,79.0);
    \node[anchor=west, font=\normalsize, color=OOrange] at (axis cs:1,79.0) [yshift=-7pt, xshift=-1em] {79.0};
    \draw[dashed, YYellow, thick] (axis cs:0,80.4) -- (axis cs:5,80.4);
    \node[anchor=west, font=\normalsize, color=YYellow] at (axis cs:1,80.4) [yshift=-7pt, xshift=-1em] {80.4};
\end{axis}

\begin{axis}[
    myaxis style,
    name=plot3,
    at={(plot2.north east)},
    anchor=north west,
    xshift=\plotHorizontalGap,
    width=0.42\textwidth,
    nodes near coords,
    nodes near coords align={south},
    nodes near coords style={text=black, font=\small, /pgf/number format/fixed, /pgf/number format/fixed zerofill, /pgf/number format/precision=1},
    legend style={
        at={(axis cs:2.5,75.5)},
        anchor=south,
        draw=none,
        legend columns=1,
    },
    xmin=1, xmax=3,
    xtick={1,2,3},
    xticklabels={0.1,\textbf{0.2},0.3},
    xlabel={(c) PRM Score Improvement @N},
]
    \addplot [sharp plot, PPink, line width=1.5pt, mark=pentagon*, mark size=2pt, mark options={fill=white,draw=PPink,line width=1pt}]
    coordinates {
    (1,79.6)(2,78.0)(3,77.9)
    };
    \addlegendentry{@1}

    \addplot [sharp plot, BBlue, line width=1.5pt, mark=pentagon*, mark size=2pt, mark options={fill=white,draw=BBlue,line width=1pt}]
    coordinates {
    (1,79.6)(2,79.3)(3,79.1)
    };
    \addlegendentry{@2}
\end{axis}

\end{tikzpicture}}
    \caption{Conditions for improved step-level self-refinement with Qwen2.5-7B-Instruct on the MATH500 dataset, where @$N$ represents between $N$ successive iteration rounds.}
    \label{fig_conditions}
    \vspace{-1em}
\end{figure}

\paragraph{Reasonable conditions for improved step-level self-refinement.} 
To further balance the performance and efficiency of step-level self-refinement, we employed Qwen2.5-7B-Instruct on the MATH500 dataset to investigate optimal refinement conditions. Based on prior knowledge, we considered the following four aspects:

\begin{itemize}[leftmargin=*]
    \item \textbf{Update Strategy.} ``Cover'' unconditionally replaces the current step with the refined step, while ``PRM\_Cover'' conditionally updates only if the refined step's PRM score surpasses that of the previous iteration. As presented in Figure~\ref{fig_condition_cover}, with increasing iteration rounds, ``PRM\_Cover'' leads to steady performance gains, whereas ``Cover'' causes serious performance degradation, even falling below the baseline. This indicates that unconditional reflection may prompt LLMs to incorrectly revise correct answers.
    \item \textbf{Number of Iterations.} Based on ``PRM\_Cover'' strategy, we experimented with up to $10$ iteration rounds to observe performance trends of sequential scaling. As depicted in Figure~\ref{fig_conditions}(a), reasoning performance peaked at the $5$th round and plateaued thereafter. Therefore, we set a maximum of $5$ iteration rounds as the condition.
\end{itemize}

The MATH500 dataset includes annotations for problem difficulty levels. We analyzed the sum of PRM scores for a solution corresponding to different problem difficulties and found a linear correlation between them, as illustrated in Figure~\ref{fig_verifier} in Appendix~\ref{appendix:verifier}. 
In other words, we can predict problem difficulty with the help of PRM scores and thus determine whether it is necessary or worthwhile for LLMs to self-reflect.
In this way, scaling is not performed for either overly easy or excessively difficult problems, where the former can avoid overthinking, and the latter allows the model to ``give up'' strategically, thereby improving efficiency.

\begin{itemize}[leftmargin=*]
    \item \textbf{Necessity of Reflection.} To avoid hallucinations from reflecting on correct steps, we set a PRM threshold to skip reflection on high-confidence steps. After testing various PRM thresholds in Figure~\ref{fig_conditions}(b), we conservatively adopt $0.9$ to flag correct steps that do not require self-reflection.
    \item \textbf{Worthiness of Reflection.} Given the complexity of reasoning tasks, LLMs may lack the capability to solve certain difficult problems. Therefore, we established a heuristic rule: if the PRM score improvement over $N$ consecutive iteration rounds falls below a threshold, the step is deemed unlikely to be corrected and further refinement is abandoned to improve efficiency. Based on experiments shown in Figure~\ref{fig_conditions}(c), we adopted a condition of ``PRM improvement over 2 rounds < 0.2'' ($\text{gap@}2<0.2$) as the criterion for the LLM to give up refinement for intractable steps.
\end{itemize}

In summary, we validate the effectiveness of fine-grained sequential scaling and define the following specific conditions for improved step-level self-refinement used in our Hybrid Test-Time Scaling method: a maximum of $5$ iteration rounds; results are overwritten only when PRM scores improve (PRM\_Cover); and iteration stops if the PRM score exceeds $0.9$ or if the PRM score improvement over two consecutive iteration rounds falls below $0.2$.

\subsection{Main Results for Hybrid Test-Time Scaling}
\paragraph{Hybrid Test-Time Scaling generally improves the reasoning performance of instruction-tuned LLMs.} 
To demonstrate the versatility of Hybrid Test-Time Scaling, we conducted main experiments on five instruction-tuned LLMs across 3B to 14B parameters from different families.
As shown in Table~\ref{tab_main}, all configurations of Hybrid TTS are based on conditional step-level self-refinement with the number of samples $N$ in BoN and the number of search paths $k$ in tree-search as variables.
Performance is evaluated using three metrics: Maj@k, RM@k, and Pass@k.
Our method achieves significant performance gains over baselines across all models and datasets. Peak improvements reach 28.6\% for LLaMA3.1-8B-Instruct on MATH500, 13.4\% for Qwen2.5-14B-Instruct on AIME24, and 21.5\% for Qwen2.5-3B-Instruct on GPQA Diamond. 
We also observe that Hybrid TTS effectively bridges the performance gap between model scales, enabling smaller LLMs to achieve the baseline reasoning performance of much larger ones.
Notably, the performance of Qwen2.5-3B/7B-Instruct and Gemma3-4B-IT on the GPQA Diamond dataset even exceeds that of the RL-enhanced DeepSeek-R1-Distill-Qwen-7B~\cite{deepseekai2025deepseekr1incentivizingreasoningcapability}, indicating that hybrid training-free TTS methods can effectively unlock the instruction-tuned LLM's latent reasoning potential.

\begin{table*}[t]
    \centering
    \resizebox{0.99\textwidth}{!}{%
    \resizebox{\textwidth}{!}{%
\begin{tabular}{lccccccccccccccccc}
\toprule[1.1pt]
 \multirow{2}{*}{\textbf{Models}} &  \textbf{Best-of-1} & \multicolumn{3}{c}{\textbf{Best-of-4}}   &\multicolumn{3}{c}{\textbf{Best-of-8}}  & \multicolumn{3}{c}{\textbf{Best-of-16}}    \\ \cmidrule(l){2-2} \cmidrule(l){3-5} \cmidrule(l){6-8}  \cmidrule(l){9-11}  
 &\multicolumn{1}{c}{Baseline} & \multicolumn{1}{c}{\texttt{Maj@4}} & \multicolumn{1}{c}{\texttt{RM@4}} & \multicolumn{1}{c}{\texttt{Pass@4}} & \multicolumn{1}{c}{\texttt{Maj@8}} & \multicolumn{1}{c}{\texttt{RM@8}} & \multicolumn{1}{c}{\texttt{Pass@8}} & \multicolumn{1}{c}{\texttt{Maj@16}} & \multicolumn{1}{c}{\texttt{RM@16}} & \multicolumn{1}{c}{\texttt{Pass@16}} \\ \midrule
\rowcolor{gray!25}  \multicolumn{11}{c}{\textit{MATH500}}  \\ 
 \multirow{1}{*}{Qwen2.5-3B}         &67.0  & 79.2   &80.6   & 83.2  & 81.4   &82.8     & 85.0   & 83.8    &84.8    &87.0      \\ 
\multirow{1}{*}{Qwen2.5-7B}   &75.2  & 84.0   &84.8   & 85.4  & 84.2   &85.4     & 86.6   &85.6     &86.2    &88.2       \\
\multirow{1}{*}{Qwen2.5-14B}   &80.8  & 85.8   &86.6   & 87.4  & 87.6   &87.8     & 88.6   &88.0     &87.8    &89.6     \\
\multirow{1}{*}{LLaMA3.1-8B} &47.2  & 64.2   &66.6   &68.4  &69.4   &71.4    & 74.2   &71.4     &73.0    &75.8       \\
\multirow{1}{*}{Gemma3-4B}   &75.6  & 80.4   &80.6   &82.4  &82.4   &82.6     & 84.2   &85.0     &84.0    &87.0         \\ \midrule
\rowcolor{gray!25}  \multicolumn{11}{c}{\textit{AIME24}}  \\
\multirow{1}{*}{Qwen2.5-3B}  &6.7  &10.0   & 16.7  &16.7  &16.7   & 13.3    & 16.7  &16.7    & 16.7   &16.7    \\ 
\multirow{1}{*}{Qwen2.5-7B}          &10.0  &20.0   &20.0   &23.3  &23.3   &23.3     & 23.3  &  16.7  &  20.0  & 20.0 \\  
\multirow{1}{*}{Qwen2.5-14B}  &13.3  &23.3   &26.7   &26.7  &23.3   &26.7     & 30.0  &  30.0  &  26.7  & 36.7 \\ 
\multirow{1}{*}{LLaMA3.1-8B} &10.0   &10.0   &10.0   &10.0  &10.0   &10.0   &16.7   &16.7 &13.3    &20.0       \\
\multirow{1}{*}{Gemma3-4B} &10.0  &13.3   &13.3   &13.3  &16.7   &16.7     &16.7  &  16.7  & 16.7 & 16.7 \\ \midrule
\rowcolor{gray!25}  \multicolumn{11}{c}{\textit{GPQA Diamond}}  \\
\multirow{1}{*}{Qwen2.5-3B}     &30.3  &35.4   &34.3   &43.9  &36.9   &39.9     & 48.0  &39.9 &41.4    &51.5 \\ 
\multirow{1}{*}{Qwen2.5-7B}        &36.4  &41.4   &40.9   &45.5  &45.5   &43.9     & 51.0  &44.4    &43.9  &51.0\\  
\multirow{1}{*}{Qwen2.5-14B}        &45.5  &51.0  &47.0   &58.1  &49.0   & 51.5    & 56.6  & 50.0    &45.5  &59.1\\ 
\multirow{1}{*}{LLaMA3.1-8B} &30.4   &30.8   &30.8  &42.9  &30.8   &32.3   &45.5   &34.3 &28.3    &41.4       \\
\multirow{1}{*}{Gemma3-4B}       &30.8  &33.3   &29.8   &39.9  &37.9   & 35.4    & 47.5 &37.9    &32.8  &53.5 \\ 
\bottomrule[1.1pt]
\end{tabular}
}}
    \caption{Main results of Hybrid Test-Time Scaling with instruction-tuned LLMs. Baseline represents their basic reasoning ability without test-time scaling, while other settings combine Best-of-N sampling and conditional step-level self-refinement in k-path tree search.}
    \label{tab_main}
    \vspace{-1em}
\end{table*}

\paragraph{Verification matters in smaller LLM test-time scaling.} 
A comparison of performance metrics reveals a consistent pattern: Maj@k and RM@k scores are generally lower than the Pass@k score. 
This discrepancy is particularly pronounced in smaller-scale LLMs, reaching a gap of up to 20.7\% for Gemma3-4B-IT on the GPQA Diamond dataset.
This indicates that while smaller models possess strong latent reasoning capabilities (high Pass@k), their ability to consistently identify and select the best answer from a set of candidates is limited (lower Maj@k and RM@k).
Therefore, to bridge this gap between latent potential and actualized performance, an accurate verification mechanism is vital—not only for final answer selection, but also for guiding the reasoning process toward the correct direction at each intermediate step.

\paragraph{High-quality verification plays a pivotal role in TTS performance.}
It is worth noting that a similar principle applies in reinforcement learning (RL), which also relies on a verifier (\emph{e.g.}, reward model) to provide accurate feedback as supervision signals. 
To empirically validate the importance of verification quality, we present a comparative analysis using two PRMs of different capabilities: the state-of-the-art Qwen2.5-Math-PRM-7B and the weaker Math-Shepherd-Mistral-7B-PRM~\cite{wang2023math}.
As shown in Table~\ref{tab_ablation_verifier}, while the performance guided by Math-Shepherd surpasses the baseline (75.2), it still lags significantly behind the results obtained with Qwen2.5-Math-PRM-7B.
This outcome is consistent with our claim in Section~\ref{sec:Verifiers} that an accurate verification is critical for maximizing reasoning performance.

\begin{table*}[t]
    \centering
    \resizebox{0.99\textwidth}{!}{%
    \begin{tabular}{l ccc ccc ccc}
\toprule
\multirow{2}{*}{\textbf{PRMs}} & \multicolumn{3}{c}{\textbf{Best-of-4}} & \multicolumn{3}{c}{\textbf{Best-of-8}} & \multicolumn{3}{c}{\textbf{Best-of-16}} \\
\cmidrule(lr){2-4} \cmidrule(lr){5-7} \cmidrule(lr){8-10}
& \texttt{Maj@4} & \texttt{RM@4} & \texttt{Pass@4} & \texttt{Maj@8} & \texttt{RM@8} & \texttt{Pass@8} & \texttt{Maj@16} & \texttt{RM@16} & \texttt{Pass@16} \\
\midrule
Math-Shepherd-Mistral-7B-PRM & 77.6 & 75.6 & 79.2 & 77.4 & 75.8 & 80.0 & 77.6 & 76.0 & 79.8 \\
Qwen2.5-Math-PRM-7B & 84.0 & 84.8 & 85.4 & 84.2 & 85.4 & 86.6 & 85.6 & 86.2 & 88.2 \\
\bottomrule
\end{tabular}}
    \caption{Comparative experiments of Hybrid Test-Time Scaling with different PRMs using Qwen2.5-7B-Instruct on the MATH500 dataset.}
    \label{tab_ablation_verifier}
    \vspace{-0.5em}
\end{table*}

\paragraph{The essence of Test-Time Scaling lies in the extensive exploration and precise exploitation of the LLM's solution space.} 
This principle is demonstrated by observing how reasoning performance varies with an increasing number of samples and search paths.
We find that all models exhibit a performance growth trend on the MATH500 dataset, suggesting that a comprehensive exploration of the solution space can enhance model reasoning capabilities.
However, results on the GPQA dataset concurrently indicate that performance gains from such an expanded search space are unstable. 
For instance, when conducting multiple experiments under identical settings, the outcomes sometimes varied significantly, as shown in Figure~\ref{fig_instability} in Appendix~\ref{appendix:instability}.
We attribute this to the inherent instability of model generation at higher temperature settings.
Nevertheless, their unexpectedly high accuracy scores (\emph{e.g.}, 59.1 and 53.5) confirm that models can possess considerable reasoning potential even without continued training.
This highlights the need for accurate external verification signals to enable more precise exploitation of the solution space, which is the core idea behind our Hybrid Test-Time Scaling-``\textit{best-of-the-best}''.
Put another way, if we can further improve the accuracy of the verifier, it is highly plausible that more consistent performance improvements could be achieved solely through training-free methods.

\subsection{Ablation Study}
\begin{table*}[t]
    \centering
    \resizebox{0.99\textwidth}{!}{%
    \resizebox{\textwidth}{!}{%
\begin{tabular}{lrllllllllllllllll}
\toprule[1.1pt]
  \multirow{2}{*}{\textbf{Methods}}  & \multicolumn{3}{c}{\textbf{Best-of-4}}   &\multicolumn{3}{c}{\textbf{Best-of-8}}  & \multicolumn{3}{c}{\textbf{Best-of-16}}    \\ \cmidrule(l){2-4} \cmidrule(l){5-7}  \cmidrule(l){8-10}  
  & \multicolumn{1}{c}{\texttt{Maj@4}} & \multicolumn{1}{c}{\texttt{RM@4}} & \multicolumn{1}{c}{\texttt{Pass@4}} & \multicolumn{1}{c}{\texttt{Maj@8}} & \multicolumn{1}{c}{\texttt{RM@8}} & \multicolumn{1}{c}{\texttt{Pass@8}} & \multicolumn{1}{c}{\texttt{Maj@16}} & \multicolumn{1}{c}{\texttt{RM@16}} & \multicolumn{1}{c}{\texttt{Pass@16}} \\ \midrule
\rowcolor{gray!25}  \multicolumn{10}{c}{\textit{Qwen2.5-3B + GPQA Diamond}}  \\
\multirow{1}{*}{BoN+Self-Refinement}      &\multicolumn{3}{c}{36.4\(_{\texttt{Pass@1}}\)}     & \multicolumn{3}{c}{37.9\(_{\texttt{Pass@1}}\)}      &\multicolumn{3}{c}{40.4\(_{\texttt{Pass@1}}\)} \\ 
\multirow{1}{*}{MCTS+BoN}    &33.8   &35.4   &\underline{40.9} &40.4   &35.9     &\underline{44.4}  &36.9 &41.4    &\underline{49.0} \\ 
\multirow{1}{*}{OpenR~\cite{wang2410openr}}    &30.3   &30.8   &37.4  &29.8  &32.3     &43.9  &33.3  &34.8     &47.0  \\ 
\multirow{1}{*}{\textbf{Hybrid Test-Time Scaling}}    &35.4   &34.3   &\textbf{43.9}  &36.9   &39.9     & \textbf{48.0}  &39.9 &41.4    &\textbf{51.5} \\ \midrule
\rowcolor{gray!25} \multicolumn{10}{c}{\textit{LLaMA3.1-8B + MATH500}}  \\
\multirow{1}{*}{BoN+Self-Refinement}     &\multicolumn{3}{c}{62.4\(_{\texttt{Pass@1}}\)}    &\multicolumn{3}{c}{66.8\(_{\texttt{Pass@1}}\)}     &\multicolumn{3}{c}{66.6\(_{\texttt{Pass@1}}\)} \\ 
\multirow{1}{*}{MCTS+BoN}    &62.2   &63.6   &\underline{64.8}  &65.2  &68.8     & \underline{71.6}  &69.6 &70.6    &\underline{73.2}\\ 
\multirow{1}{*}{OpenR~\cite{wang2410openr}}    &48.6   &47.6   &54.2  &60.2    &60.2     &63.0  & 60.4   & 60.0    & 61.2 \\ 
\multirow{1}{*}{\textbf{Hybrid Test-Time Scaling}}   & 64.2   &66.6   &\textbf{68.4}  &69.4   &71.4    & \textbf{74.2}   &71.4     &73.0    &\textbf{75.8}       \\ \midrule
\rowcolor{gray!25} \multicolumn{10}{c}{\textit{Qwen2.5-14B + AIME24}}  \\
\multirow{1}{*}{BoN+Self-Refinement}     &\multicolumn{3}{c}{13.3\(_{\texttt{Pass@1}}\)}    &\multicolumn{3}{c}{16.7\(_{\texttt{Pass@1}}\)}     &\multicolumn{3}{c}{20.0\(_{\texttt{Pass@1}}\)} \\ 
\multirow{1}{*}{MCTS+BoN}    &13.3   &16.7   &16.7  &16.7  &23.3     & \underline{23.3}  &20.0 &23.3    &\underline{26.7}\\ 
\multirow{1}{*}{OpenR~\cite{wang2410openr}}    &16.7   &20.0   &\underline{20.0}  &16.7    &16.7     &16.7  & 23.3   & 23.3    & 23.3 \\ 
\multirow{1}{*}{\textbf{Hybrid Test-Time Scaling}}   & 23.3   &26.7   &\textbf{26.7}  &23.3   & 26.7    & \textbf{30.0}   &30.0     &26.7    &\textbf{36.7}       \\
\bottomrule[1.1pt]
\end{tabular}
}}
    \caption{Ablation study on pairwise combinations from our Hybrid Test-Time Scaling methods, where the best results are shown in \textbf{bold} and the second-best results are \underline{underlined}. ``BoN+Self-Refinement'' represents one reasoning path for every question, so only \texttt{Pass@1} is reported. ``MCTS+BoN'' is parallel-only scaling, and ``OpenR'' is solution-level Hybrid Test-Time Scaling.}
    \label{tab_ablation}
    \vspace{-1em}
\end{table*}
To validate the necessity of integrating all three components of our method, we conducted ablation studies on their pairwise combinations.
The results across various models and datasets are summarized in Table~\ref{tab_ablation}.
To ensure a fair comparison, all tested combinations employ the same PRM verifier and operate at the step level. 
We also compare our method against OpenR~\cite{wang2410openr}, a solution-level baseline that combines MCTS with multi-round self-refinement under the identical refinement conditions, where the self-refinement serves as the MCTS simulation.
The results indicate that while all pairwise combinations improve upon the baseline, none match the performance of our complete three-component hybrid strategy. 
This finding strongly suggests that each component contributes meaningfully to the overall performance, providing compelling support for our conclusion that a fine-grained hybrid inference paradigm yields superior performance.

A detailed analysis of the computational overhead for these methods is provided in Appendix~\ref{appendix:overhead}. In practice, a maximum of $5$ iteration rounds and relatively conservative stopping conditions inevitably increases inference cost. However, in order to fully explore the upper bound of the reasoning capabilities of training-free TTS, we chose a strategy that maximizes reasoning performance.

\section{Conclusion}
In this paper, we propose Step-level Verifier-guided Hybrid Test-Time Scaling for LLM reasoning.
First, we introduce a conditional step-level self-refinement method to balance performance and efficiency.
Building on this, we synergistically combine this fine-grained sequential scaling method with representative parallel scaling techniques (\emph{e.g.}, Best-of-N and MCTS) at the step level.
The effectiveness of this hybrid inference paradigm demonstrates that different sequential and parallel scaling methods can be orthogonal and complementary.
Finally, our large-scale experiments corroborate that the essence of test-time scaling is to deeply explore and precisely exploit the LLM's solution space.
The success in enhancing smaller instruction-tuned LLMs via training-free TTS methods not only establishes the pivotal role of accurate step-level verifiers but also inspires a new direction for training-based TTS with smaller LLMs.

\section*{Limitations}
We only perform inference once in the experiments, so there may be some performance fluctuations.
However, the purpose of this paper is to explore the reasoning performance boundaries achievable by hybrid training-free TTS methods.
This conclusion has already been confirmed by most experiments, and training-free TTS methods even have the potential to outperform training-based TTS methods.

\section*{Acknowledgments}
This work was supported in part by the National Science Foundation of China (Nos. 62276056 and U24A20334), the Yunnan Fundamental Research Projects (No. 202401BC070021), the Yunnan Science and Technology Major Project (No. 202502AD080014), and the Program of Introducing Talents of Discipline to Universities, Plan 111 (No.B16009).

\bibliography{Hybrid_TTS}
\onecolumn
\clearpage
\appendix
\section{Meta-prompts for Self-refinement}
\label{appendix:meta-prompt}
Following the principles of efficient prompting outlined in~\citet{chang2024efficient}, we designed meta-prompts for both step- and solution-level self-refinement to achieve a performance-efficiency trade-off.
\begin{tcolorbox}[width=\textwidth, colback=blue!5!white, colframe=blue!75!black, title=Prompt for step-level self-refinement]
\noindent
\textbf{CoT Prompt:} Please reason step by step, and put your final answer within boxed\{{}\}.\\
\\
\textbf{Critic Prompt:} There is a weak reasoning step in the solution, please provide a strict reflection to correct only this one step with less than 150 tokens. Don't output the complete solution.\\
\\
\textbf{Rewrite Prompt:} Please refine the weak answer according to your Reflection. Not allowed to use code to solve the question. Please reason step by step, and put your final answer within boxed\{{}\}.
\end{tcolorbox}

\begin{tcolorbox}[width=\textwidth, colback=blue!5!white, colframe=blue!75!black, title=Prompt for solution-level self-refinement]
\noindent
\textbf{CoT Prompt:} Please reason step by step, and put your final answer within boxed\{{}\}.\\
\\
\textbf{Critic Prompt:} There is a weak solution, please provide a strict reflection to correct it.\\
\\
\textbf{Rewrite Prompt:} Please refine the weak answer according to your Reflection. Not allowed to use code to solve the question. Please reason step by step, and put your final answer within boxed\{{}\}.
\end{tcolorbox}

\begin{tcolorbox}[width=\textwidth, colback=blue!5!white, colframe=blue!75!black, title=Prompt for step-level self-refinement (GPQA)]
\noindent
\textbf{CoT Prompt:} Please reason step by step, and put your final answer within boxed\{{LETTER}\}(without quotes) where LETTER is one of ABCD.\\
\\
\textbf{Critic Prompt:} There is a weak reasoning step in the solution, please provide a strict reflection to correct only this one step with less than 150 tokens. Don't output the complete solution.\\
\\
\textbf{Rewrite Prompt:} Please refine the weak answer according to your Reflection. Not allowed to use code to solve the question. Please reason step by step, and put your final answer within boxed\{{LETTER}\}(without quotes) where LETTER is one of ABCD.
\end{tcolorbox}

\begin{tcolorbox}[width=\textwidth, colback=blue!5!white, colframe=blue!75!black, title=Prompt for solution-level self-refinement (GPQA)]
\noindent
\textbf{CoT Prompt:} Please reason step by step, and put your final answer within boxed\{{LETTER}\}(without quotes) where LETTER is one of ABCD.\\
\\
\textbf{Critic Prompt:} There is a weak solution, please provide a strict reflection to correct it.\\
\\
\textbf{Rewrite Prompt:} Please refine the weak answer according to your Reflection. Not allowed to use code to solve the question. Please reason step by step, and put your final answer within boxed\{{LETTER}\}(without quotes) where LETTER is one of ABCD.
\end{tcolorbox}

\section{The Algorithm for Hybrid Test-Time Scaling}
\label{appendix:algorithm}
We combine our proposed conditional step-level self-refinement with two classical parallel scaling methods, Best-of-N and Tree Search, under the guidance of a step-level verifier, and design an algorithm to describe the hybrid strategy.

\begin{algorithm*}
    \caption{Hybrid Test-Time Scaling}
    \label{al:Hybrid_Scaling}
    \begin{algorithmic}[1]
        \State \textbf{Input:} Problem $P$, search paths $N_p$, samples $N_s$, self-refinement iterations $N_r$, LLM, PRM
        \State \textbf{Output:} Final Solution $s$

        \State Initialize root node $n_0$ from $P$

        \For{$i = 1$ to $N_p$} \Comment{\textbf{Tree Search}}
            \State Initialize $n_{current} \gets n_0$, $p_i \gets \text{""}$

            \While{<eos> is not in $n_{current}.text$}
                \State Generate $N_s$ candidate steps $\mathcal{S}$ using $\text{LLM}(P, p_i)$ \Comment{Sampling}
                \State Evaluate scores $\mathcal{V}$ for steps in $\mathcal{S}$ using $\text{PRM}(P, p_i, \cdot)$ to calculate P-UCB \Comment{Evaluation}
                \State Select $s_{best}$ from $\mathcal{S}$ based on P-UCB as next step $n_{next}.text$ \Comment{\textbf{Best-of-N}}

                \For{$j = 1$ to $N_r$} 
                    \If{IsRefinement($n_{next}.text$)} \Comment{\textbf{Conditional Self-refinement}}
                       \State $C \gets \text{LLM}(\texttt{CritiquePrompt}(P, p_i, n_{next}.text))$ \Comment{Reflection}
                       \State $r \gets \text{LLM}(\texttt{RefinePrompt}(P, p_i, n_{next}.text, C))$ \Comment{Revision}
                       \If{$\text{PRM}(P, p_i, r) > \text{PRM}(P, p_i, n_{next}.text)$}
                           \State $n_{next}.text \gets r$
                       \EndIf
                    \EndIf
                \EndFor

                \State $p_i \gets \text{Concatenate}(p_i, n_{next}.text)$
            \EndWhile

            \State $\mathcal{P} \gets \mathcal{P} \cup \{p_i\}$
        \EndFor

        \State $s \gets \text{Majority Voting / Reward Model Scoring}(\mathcal{P})$
        \State \Return $s$.
    \end{algorithmic}
\end{algorithm*}

\begin{table*}[h]
    \centering
    \resizebox{0.99\textwidth}{!}{%
    \resizebox{\textwidth}{!}{%
\begin{tabular}{l ccc ccc ccc}
\toprule[1.1pt]
 \multirow{2}{*}{\textbf{Methods}}  & \multicolumn{3}{c}{\textbf{Best-of-4}}    & \multicolumn{3}{c}{\textbf{Best-of-8}}    & \multicolumn{3}{c}{\textbf{Best-of-16}}   \\ \cmidrule(lr){2-4} \cmidrule(lr){5-7}  \cmidrule(lr){8-10}
  & \multicolumn{1}{c}{FLOPs} & \multicolumn{1}{c}{\# Tokens} & \multicolumn{1}{c}{Acc.} & \multicolumn{1}{c}{FLOPs} & \multicolumn{1}{c}{\# Tokens} & \multicolumn{1}{c}{Acc.} & \multicolumn{1}{c}{FLOPs} & \multicolumn{1}{c}{\# Tokens} & \multicolumn{1}{c}{Acc.} \\ \midrule
\rowcolor{gray!20}  \multicolumn{10}{c}{\textit{Qwen2.5-3B + GPQA Diamond}}  \\
BoN+Self-Refinement        & 7.41$\times 10^9$& 9580.23  & 36.4 & 8.12$\times 10^9$& 14422.07 & 37.9 & \textbf{9.33$\times 10^9$} & \textbf{22583.35} & \textbf{40.4} \\
MCTS+BoN                   & 7.37$\times 10^9$& 9276.94  & 40.9 & \textbf{10.34$\times 10^9$} & \textbf{29424.34} & \textbf{44.4} & 17.80$\times 10^9$& 80049.37 & 49.0 \\
Hybrid Test-Time Scaling   & \textbf{9.69$\times 10^9$} & \textbf{25006.00} & \textbf{43.9} & 14.25$\times 10^9$& 55957.14 & 48.0 & 22.62$\times 10^9$& 112728.73 & 51.5 \\ \midrule
\rowcolor{gray!20} \multicolumn{10}{c}{\textit{Qwen2.5-14B + AIME24}}  \\
BoN+Self-Refinement        & 34.72$\times 10^9$& 16418.77 & 13.3 & 36.29$\times 10^9$& 20230.50 & 16.7 & \textbf{51.83$\times 10^9$} & \textbf{58182.17} & \textbf{20.0} \\
MCTS+BoN                   & 33.21$\times 10^9$& 12723.53 & 16.7 & \textbf{47.95$\times 10^9$} & \textbf{48700.07} & \textbf{23.3} & 91.48$\times 10^9$& 154984.13 & 26.7 \\
Hybrid Test-Time Scaling   & \textbf{50.20$\times 10^9$} & \textbf{54189.77} & \textbf{26.7} & 122.79$\times 10^9$& 231410.97 & 30.0 & 150.21$\times 10^9$& 298355.30 & 36.7 \\
\bottomrule[1.1pt]
\end{tabular}
}}
    \caption{The computational overhead of different combinations of training-free TTS methods.}
    \label{tab_computation}
    \vspace{-1em}
\end{table*}

\twocolumn

\section{The Computational Overhead of Different Combinations of TTS Methods}
\label{appendix:overhead}
As stated in~\citet{wu2024inference}, the inference FLOPs can be estimated following the Eq.\ref{FLOPs} defined in~\citet{kaplan2020scaling}:
\begin{equation}
    \label{FLOPs}
    \text{FLOPs}(N,T;S) \approx 2N + 2n_{\text{layer}} \times T \times d_{\text{model}}
\end{equation}
where FLOPs depends on the number of model parameters $N$, the number of generated tokens $T$ and the model architecture settings $n_{\text{layer}}$ and $d_{\text{model}}$.

To fairly compare the cost-effectiveness across parallel, sequential, and hybrid methods under equal compute budgets, we consider fixed model configurations.
In this setting, the inference FLOPs are primarily determined by the total number of generated tokens, which is aggregated from all generative steps within our framework, primarily from:
\begin{itemize}
    \item Best-of-N (BoN): Sampling multiple step candidates based on the question and context.
    \item Tree Search (MCTS): Expanding nodes (steps) based on the question and context.
    \item Self-refinement: Generating an initial step-wise response based on the question and context, reflecting on potentially erroneous steps, and rewriting them based on the feedback.
\end{itemize}

Table~\ref{tab_computation} presents the average Inference FLOPs per question, average generation length, and performance for Qwen2.5-3B/14B-Instruct on the GPQA Diamond and AIME24 benchmarks, respectively.
From these results, we observe that under comparable computational costs (\emph{e.g.}, Best-of-16 for BoN+Self-Refinement, Best-of-8 for MCTS+BoN, and Best-of-4 for Hybrid TTS), Hybrid Test-Time Scaling consistently achieves superior performance. 
In contrast, using only pairwise combinations of these strategies (\emph{e.g.}, MCTS+BoN or BoN+Self-Refinement)—even with a larger sampling budget—leads to more limited performance gains.
Only the synergistic combination of all three strategies allows the model to fully realize its potential with training-free TTS. 
This demonstrates the cost-effectiveness advantage of the Hybrid Test-Time Scaling strategy.

\section{Verifier-indicated Difficulty Levels}
\label{appendix:verifier}
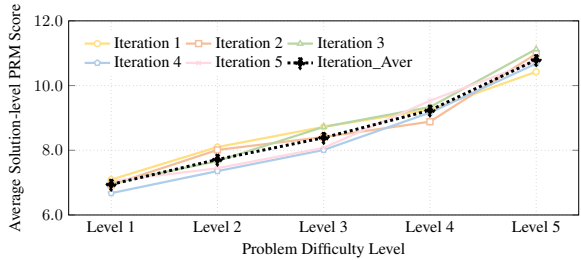
\begin{figure}
    \centering
    \resizebox{0.99\linewidth}{!}{%
    \begin{tikzpicture}
\begin{axis}[
    at={(0,0)},
    anchor=south west,
    height=0.25\textheight,
    width=0.85\textwidth,
    grid=major,
    major grid style={dotted},
    enlarge x limits=0.1,
    ymin=6, ymax=12,
    yticklabel style={/pgf/number format/fixed, /pgf/number format/fixed zerofill, /pgf/number format/precision=1},
    ylabel={Average Solution-level PRM Score}, 
    ylabel style={anchor=center},
    xtick style={draw=none},
    legend entries={Iteration 1,Iteration 2,Iteration 3,Iteration 4,Iteration 5, Iteration\_Aver},
    legend columns=3,
    legend style={
        at={(axis cs:2.3,10.3)},
        anchor=south,
        draw=none,
        legend cell align=left,
    },
    xmin=1, xmax=5,
    xtick={1,2,3,4,5},
    xticklabels={Level 1, Level 2, Level 3, Level 4, Level 5},
    xlabel={Problem Difficulty Level},
]
    \addplot [sharp plot, YYellow!50, line width=1.5pt, mark=*, mark size=2pt, mark options={fill=white,draw=YYellow!50,line width=1pt}]
    coordinates {
    (1,7.093023) (2,8.1) (3,8.72381) (4,9.2578139) (5,10.42537) 
    };
    \addplot [sharp plot, OOrange!50, line width=1.5pt, mark=square*, mark size=2pt, mark options={fill=white,draw=OOrange!50,line width=1pt}]
    coordinates {
    (1,6.930233) (2,8.011111) (3,8.409524) (4,8.882813) (5,10.98507)
    };
    \addplot [sharp plot, GGreen!50, line width=1.5pt, mark=triangle*, mark size=2pt, mark options={fill=white,draw=GGreen!50,line width=1pt}]
    coordinates {
    (1,6.976744) (2,7.666667) (3,8.72381) (4,9.335938) (5,11.12687)
    };
    \addplot [sharp plot, BBlue!50, line width=1.5pt, mark=pentagon*, mark size=2pt, mark options={fill=white,draw=BBlue!50,line width=1pt}]
    coordinates {
    (1,6.674419) (2,7.355556) (3,8.009524) (4,9.164063) (5,10.6791)
    };
    \addplot [sharp plot, PPink!50, line width=1.5pt, mark=x, mark size=2pt, mark options={fill=white,draw=PPink!50,line width=1pt}]
    coordinates {
    (1,7.046512) (2,7.444444) (3,8.07619) (4,9.53125) (5,10.76119)
    };
    \addplot [sharp plot, black, dotted, line width=2pt, mark=diamond*, mark size=3.5pt, mark options={fill=black, draw=black}]
    coordinates {
    (1,6.9441862) (2,7.7155556) (3,8.3885716) (4,9.23437558) (5,10.79552)
    };
\end{axis}
\end{tikzpicture}}
    \caption{The correlation between solution-level PRM and problem difficulty level, where the black dashed line represents the average of all iterations.}
    \label{fig_verifier}
    \vspace{-1em}
\end{figure}
We apply the ``PRM\_Cover'' update strategy with Qwen2.5-7B-Instruct for five iterations on the MATH500 dataset.
In each iteration, we calculate the sum of PRM scores for each solution corresponding to a problem.
Then, we aggregate and average the PRM scores for problems of the same difficulty level, using this average solution-level PRM score to represent the difficulty of that category of problems.
As shown in Figure~\ref{fig_verifier}, we found a linear correlation between the solution-level PRM and the problem difficulty level.

\section{Instability of LLM during Training-free Test-Time Scaling}
\label{appendix:instability}
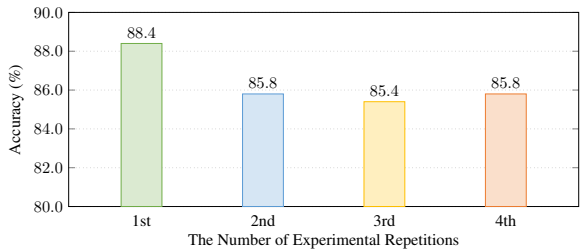
\begin{figure}
    \centering
    \resizebox{0.99\linewidth}{!}{%
    \begin{tikzpicture}
    \begin{axis}[
        at={(0,0)},
        anchor=south west,
        height=0.25\textheight,
        width=0.85\textwidth,
        ymajorgrids,
        grid style=dotted,
        ybar,
        bar width=2.5em,
        bar shift=0pt,
        enlarge x limits=0.2,
        nodes near coords,
        nodes near coords align={vertical},
        nodes near coords style={/pgf/number format/fixed, /pgf/number format/fixed zerofill, /pgf/number format/precision=1},
        ylabel={Accuracy (\%)},
        xlabel={The Number of Experimental Repetitions},
        symbolic x coords={1st,2nd,3rd,4th},
        xtick={1st,2nd,3rd,4th},
        x tick label style={inner sep=1pt},
        xtick style={draw=none},
        ymin=80, ymax=90,
        yticklabel style={/pgf/number format/fixed, /pgf/number format/fixed zerofill, /pgf/number format/precision=1},
    ]
        \addplot [fill=GGreen!25, draw=GGreen] coordinates {(1st, 88.4)};
        \addplot [fill=BBlue!25, draw=BBlue] coordinates {(2nd, 85.8)};
        \addplot [fill=YYellow!25, draw=YYellow] coordinates {(3rd, 85.4)};
        \addplot [fill=OOrange!25, draw=OOrange] coordinates {(4th, 85.8)};
    \end{axis}
\end{tikzpicture}}
    \caption{Results of repeated trials under exactly the same experimental setup.}
    \label{fig_instability}
    \vspace{-1em}
\end{figure}
In our experiments, we typically perform inference only once for a set of setting. 
However, during the experimental process, we found that the model's performance can vary significantly under the same setting.
Figure~\ref{fig_instability} shows the results of the Qwen2.5-7B-Instruct on the MATH500 dataset with the setting of Best-of-4 and 16-path search.

\section{Case Study of Step-level and Solution-level Self-refinement}
Qwen2.5-7B-Instruct performed 5 rounds of step- and solution-level self-refinement on its initial generation for the same problem. The solution-level self-refinement failed to correct the error through iteration, whereas step-level self-refinement corrected the error in \textbf{Step3} after the first round, with its PRM score improving from 0.0596 to 0.8711.

\begin{tcolorbox}[
    title=Case Study,
    width=\textwidth,
    colback=gray!5, 
    colframe=gray!75!black,
    fonttitle=\bfseries
]
\begin{description}[leftmargin=1cm]
    \item[Question:] 
    The binary number $10101001110_{2}$ is equal to what number in base eight?

    \item[Ground Truth:]
    Since $2^3=8$, we may convert between base 2 and base 8 representations by replacing each block of three digits in base 2 with its equivalent in base 8. In this case, we begin by noticing that the last three digits are worth $110_2=6_8$. The next block of three digits is $001_2=1_8$. Continuing, we find that the next two digits (moving right-to-left) are $101_2=5_8$ and $010_2=2_8$. Altogether, we find that $10101001110_{2}=\boxed{2516_8}$.
    
    \item[Initial Response:]
    To convert the binary number $10101001110_{2}$ to an octal (base eight) number, we can follow these steps:
    \begin{enumerate}
        \item \textbf{Group the binary digits into sets of three, starting from the right:} $10101001110_{2} \rightarrow 101 \, 010 \, 011 \, 110$
        \item \textbf{Convert each group of three binary digits to its octal equivalent:}
              \begin{itemize}
                  \item $101_{2} = 1 \cdot 2^2 + 0 \cdot 2^1 + 1 \cdot 2^0 = 4 + 0 + 1 = 5_{8}$
                  \item $010_{2} = 0 \cdot 2^2 + 1 \cdot 2^1 + 0 \cdot 2^0 = 0 + 2 + 0 = 2_{8}$
                  \item $011_{2} = 0 \cdot 2^2 + 1 \cdot 2^1 + 1 \cdot 2^0 = 0 + 2 + 1 = 3_{8}$
                  \item $110_{2} = 1 \cdot 2^2 + 1 \cdot 2^1 + 0 \cdot 2^0 = 4 + 2 + 0 = 6_{8}$
              \end{itemize}
        \item \textbf{Combine the octal digits:} $10101001110_{2} = 5236_{8} $
        
        Thus, the binary number $10101001110_{2}$ is equal to $\boxed{5236_{8}}$.\quad (\textcolor{red}{Incorrect})
    \end{enumerate}
    \item[Solution-Level:]
    Let's convert the binary number $10101001110_2$ to octal equivalent step by step.
    \begin{enumerate}
        \item \textbf{Group the binary digits into sets of three, starting from the right:} $10101001110_2 \rightarrow 101 \, 010 \, 011 \, 110$
        \item \textbf{Convert each group of three binary digits to its octal equivalent:}
              \begin{itemize}
                  \item For $101_2$: $1 \cdot 2^2 + 0 \cdot 2^1 + 1 \cdot 2^0 = 4 + 0 + 1 = 5_8$
                  \item For $010_2$: $0 \cdot 2^2 + 1 \cdot 2^1 + 0 \cdot 2^0 = 0 + 2 + 0 = 2_8$
                  \item For $011_2$: $0 \cdot 2^2 + 1 \cdot 2^1 + 1 \cdot 2^0 = 0 + 2 + 1 = 3_8$
                  \item For $110_2$: $1 \cdot 2^2 + 1 \cdot 2^1 + 0 \cdot 2^0 = 4 + 2 + 0 = 6_8$
              \end{itemize}
        \item \textbf{Combine the octal digits:} $10101001110_2 = 5236_8$
    \end{enumerate}
    Thus, the binary number $10101001110_2$ is equal to $\boxed{5236_8}$.\quad (\textcolor{red}{Incorrect})
    
    \item[Initial Step3:] $10101001110_2 \rightarrow 101 \, 010 \, 011 \, 110$

    \item[Refined Step3:] $10101001110_2 \rightarrow 010 \, 101 \, 001 \, 110$

   \item[Step-Level:]
    Let's convert the binary number $10101001110_2$ to base eight step by step.
    \begin{enumerate}
        \item \textbf{Group the binary digits into sets of three, starting from the right:} $10101001110_2 \rightarrow 010 \, 101 \, 001 \, 110$. We added a leading zero to the leftmost group to make it a set of three digits.
        \item \textbf{Convert each group of three binary digits to its equivalent octal digit:}
              \begin{itemize}
                  \item $010_2 = 2_8$
                  \item $101_2 = 5_8$
                  \item $001_2 = 1_8$
                  \item $110_2 = 6_8$
              \end{itemize}

        \item \textbf{Combine the octal digits to form the final octal number:} $010 \, 101 \, 001 \, 110_2 \rightarrow 2516_8$ 
    \end{enumerate}
    Therefore, the binary number $10101001110_2$ is equal to $\boxed{2516_8}$.\quad (\textcolor{green}{Correct})

\end{description}

\end{tcolorbox}

\end{document}